\documentclass[conference]{IEEEtran}
\IEEEoverridecommandlockouts
\usepackage{cite}
\usepackage{amsmath,amssymb,amsfonts}
\usepackage{algorithmic}
\usepackage{graphicx}
\usepackage{booktabs} 
\usepackage{longtable}
\usepackage{supertabular}
\usepackage{easyReview}
\usepackage{listings}
\usepackage{textcomp}
\usepackage{xcolor}
\usepackage{pythonhighlight}
\usepackage{multicol}
\usepackage{tabularray}  
\usepackage{amssymb}
\usepackage{lipsum}
\usepackage[ruled,vlined]{algorithm2e}
\usepackage{amsmath}
\usepackage{multirow}
\usepackage{easyReview}
\makeatletter
\newcommand{\linebreakand}{%
  \end{@IEEEauthorhalign}
  \hfill\mbox{}\par
  \mbox{}\hfill\begin{@IEEEauthorhalign}
}
\makeatother
\def\BibTeX{{\rm B\kern-.05em{\sc i\kern-.025em b}\kern-.08em
T\kern-.1667em\lower.7ex\hbox{E}\kern-.125emX}}
\usepackage{todonotes}

\begin{document}
\title{
Contextual Embedding-based Clustering to Identify Topics for Healthcare Service Improvement

% \thanks{Funding Info*}
}
%\IEEEoverridecommandlockouts
%\IEEEpubid{\begin{minipage}[t]{\textwidth}\ \\[10pt]
%\centering\normalsize{copyright-placeholder}
%\end{minipage}}

\author{\IEEEauthorblockN{K M Sajjadul Islam}
\IEEEauthorblockA{\textit{Marquette University} \\
% \textit{Marquette University}\\
sajjad.islam@marquette.edu}
\and
\IEEEauthorblockN{Ravi Teja Karri}
\IEEEauthorblockA{\textit{Froedtert Health, Inc} \\
% textit{Froedtert Health, Inc}\\
%Platteville, WI, USA \\
raviteja.karri@froedtert.com}
\and
\IEEEauthorblockN{Srujan Vegesna}
\IEEEauthorblockA{\textit{Froedtert Health, Inc} \\
% \textit{Name of the University}\\
%Platteville, WI, USA \\
srujan.vegesna@froedtert.com}
\linebreakand
\IEEEauthorblockN{Jiawei Wu}
\IEEEauthorblockA{\textit{Medical College of Wisconsin} \\
% \textit{Name of the University}\\
%Milwaukee, WI, USA  \\
jiawu@mcw.edu}
\and
\IEEEauthorblockN{Praveen Madiraju}
\IEEEauthorblockA{\textit{Marquette University} \\
% \textit{Marquette University}\\
praveen.madiraju@marquette.edu}
}
\maketitle 

\begin{abstract}
Understanding patient feedback is crucial for improving healthcare services, yet analyzing unlabeled short-text feedback presents challenges due to limited data and domain-specific nuances. Traditional supervised approaches require extensive labeled datasets, making unsupervised methods more practical for extracting insights. This study applies unsupervised techniques to analyze 439 survey responses from a healthcare system in Wisconsin, USA. A keyword-based filter was used to isolate complaint-related feedback using a domain-specific lexicon. To identify dominant themes, we evaluated traditional topic models such as Latent Dirichlet Allocation (LDA) and Gibbs Sampling Dirichlet Multinomial Mixture (GSDMM) - alongside BERTopic, a neural embedding-based clustering method. To improve coherence and interpretability in sparse, short-text data, we propose kBERT, which integrates BERT embeddings with k-means clustering. Model performance was assessed using coherence scores ($C_v$) and average Inverted Rank-Biased Overlap ($\operatorname{IRBO_{avg}}$). kBERT achieved the highest coherence ($C_v$ = 0.53) and topic separation ($\operatorname{IRBO_{avg}}$ = 1.00), outperforming all other models. These findings highlight the value of embedding-based, context-aware models in healthcare analytics.
\end{abstract}


\section{Introduction}

Understanding and addressing complaints is essential across domains, including healthcare. Linguistic studies define a complaint as a form of communication expressing dissatisfaction when expectations are unmet \cite{olshtain198710}. While industries like e-commerce and utilities increasingly use feedback to improve services, healthcare systems are also beginning to leverage patient comments for quality enhancement. Complaint identification is often approached through supervised learning, but such methods require large labeled datasets, which are resource-intensive to build. As a result, unsupervised techniques like topic modeling and clustering have gained attention for their ability to extract meaningful insights from unlabeled text.

Topic modeling aims to identify latent themes within a collection of documents by analyzing word distributions. Methods such as Latent Dirichlet Allocation (LDA) and Gibbs Sampling Dirichlet Multinomial Mixture (GSDMM) have been widely applied to textual data to uncover dominant topics \cite{churchill2022evolution}.  Clustering methods, in contrast, group similar documents based on their feature representations. Unlike topic models, which allow words to belong to multiple topics, clustering techniques assign each document to a single group based on textual similarity. Recent advances in neural embedding-based clustering, such as BERTopic, have been explored for their potential to handle short and noisy texts compared to LDA and GSDMM. \cite{curiskis2020evaluation}. While topic modeling methods and neural embedding-based clustering approaches have been explored for various text analysis tasks, their effectiveness in scenarios where data is both short and limited remains an open question. In this study, we develop a neural embedding-based clustering model, kBERT, and systematically assess its performance alongside established methods to evaluate their suitability for analyzing short-text on limited data.

This study presents a system for analyzing patient survey feedback from hospital visits, with an emphasis on identifying complaint-related topics. To address the challenges of short, sparse, and domain-specific text, we use a curated keyword list to filter negative feedback and apply unsupervised methods to uncover dominant themes. We evaluate traditional topic models (LDA, GSDMM) and embedding-based approaches (BERTopic, kBERT) on a small short-text dataset. By combining domain-aware filtering with neural embeddings, our approach aims to improve topic coherence and interpretability in low-resource healthcare settings. This work is guided by the following research questions (RQs):

\textbf{RQ1:} How can unsupervised techniques be effectively utilized to identify and categorize topics in short-text patient feedback?

\textbf{RQ2:} What are the key challenges in analyzing healthcare-specific short-texts and how can advanced natural language processing (NLP) techniques improve extracting meaningful insights from patient feedback in a hospital setting?

\textbf{RQ3:} How can the performance of unsupervised topic modeling and clustering methods be effectively evaluated in short-text patient feedback analysis?

\section{Related Work}
\subsection{Topic Modeling and Word Embeddings-based Clustering} 
In NLP, complaint identification is often treated as a classification task when labeled data is available. Traditional approaches use machine learning models such as SVC, Random Forest, and MLP, as well as deep learning methods like CNNs. While basic sentiment analysis tools (e.g., VADER, TextBlob, NLTK) can classify feedback as positive, negative, or neutral, they often miss domain-specific nuances. Recent advancements in large language models (LLMs) have significantly improved various NLP tasks \cite{nipu2024reliable}. Frameworks like the Complaint Text Classification Framework (CCF) leverage LLM-based feature extraction to enhance classification accuracy \cite{yuan2024framework}. While classification is helpful, it often lacks deeper interpretability. Topic modeling techniques like LDA are widely used to uncover latent themes in textual data and have been applied across domains to analyze customer feedback. As the field evolves, several studies have compared topic modeling methods to assess their effectiveness. For example, Jung et al. evaluated LDA, NMF, CTM, and BERTopic, finding that BERT-based models outperformed others in both topic diversity and coherence \cite{jung2024expansive}. Embedding-based methods have significantly advanced topic modeling by utilizing contextual embeddings like BERT and SBERT to capture semantic and syntactic nuances. The Embedded Topic Model (ETM) \cite{dieng2020topic} integrates LDA with embeddings to enhance topic coherence. 
%Approaches combining BERT embeddings with clustering methods show improved coherence and interpretability. For instance, a framework integrating BERT with clustering and dimensionality reduction demonstrated superior clustering quality \cite{george2023integrated}. 

\subsection{Topic Modeling for Short-Text}
Conventional methods like LDA struggle with short texts due to their reliance on bag-of-words representations, which often fail to capture contextual nuances \cite{udupa2022exploratory}. Recent studies have shown that alternative models like GSDMM and BERTopic outperform LDA in short-text scenarios. For example, Udupa et al. demonstrated the effectiveness of both GSDMM and BERTopic for topic clustering in brief texts \cite{udupa2022exploratory}. Additionally, clustering token-level contextualized embeddings from models like BERT has proven effective in capturing polysemy and organizing documents, often matching or exceeding LDA in performance \cite{thompson2020topic}. These findings indicate a shift toward models better suited for interpreting limited-context data.

\subsection{Topic Modeling in Healthcare Feedback Service}
LLMs are increasingly being applied to healthcare-related NLP tasks \cite{manir2024llm}, while recent studies have focused on analyzing patient feedback to enhance healthcare service quality. For instance, Alexander et al. \cite{alexander2022automating} have developed a tool for data analysis and visualization that leverages the BERTopic model to identify key topics and trends in patient feedback collected from NHS services in England. This approach enables a dynamic overview of sentiment changes over time. Similarly, Osváth et al. \cite{osvath2023analyzing} have applied BERT embeddings coupled with sentiment analysis to dissect patient experiences and public reactions sourced from a Hungarian online forum. Their work aims to uncover meaningful patterns and emotional responses, which are critical for enhancing the quality of healthcare services. These studies illustrate the growing application of advanced topic modeling techniques in healthcare feedback analysis, highlighting their potential to improve service understanding and delivery. 

To best of our knowledge, no prior study has addressed short-text topic modeling on limited datasets, specifically in the context of healthcare service.

\begin{table}[tb]
  \caption{Sample Keyword (Hot Word) List}
  \label{keyword-list-table}
  \centering
  \begin{tabular}{l|lllll}
   \toprule
     Type & \multicolumn{5}{c}{Examples}\\
    \midrule
    Full & angry & malpractice & delay & grievance & african \\
    Short & negl & abus & argu & negl & Viol \\
    Jargon & hipaa &  hippa & MRSA  &  &  \\
    Prefix & un &  mis &   &  &  \\
    \bottomrule
  \end{tabular}
\end{table}

\section{Methodology}
This study analyzes short-text patient feedback using keyword filtering and unsupervised topic modeling. GSDMM serves as the baseline for short-text modeling, while LDA provides a long-text reference. We also implement BERTopic and kBERT, which leverage transformer-based embeddings to enhance coherence.

\subsection{Dataset Description}\label{sec-dataset}

Patient comments are collected from a health system in Wisconsin, USA, through surveys. This feedback offers detailed insights into patient experiences. The surveys include feedback on various aspects, such as hospital stays, emergency department visits, clinic appointments, ambulatory surgery, outpatient services, and urgent care. A total of 439 patient comments were collected for analysis and testing. To filter complaint comments, the ‘Patient Experience and Quality’ team created a list of 202 specific keywords, referred to as “hot words”, including domain-specific terms related to medical procedures, patient safety, and service quality. The dataset is private and consists of short-text feedback, with a median word count of 14.

To evaluate the robustness of our methodology, we also tested it on two public datasets: the Clickbait-title dataset \cite{chakraborty2016stop} and the 20Newsgroup dataset \cite{20NewsGroupDataset}. The Clickbait-title dataset contains short headlines labeled as clickbait or non-clickbait. We selected 10,000 samples with a median word count of 9, and also created a smaller subset of 500 samples to evaluate performance on small short-text data. The 20Newsgroup dataset consists of longer articles across 20 distinct topics and is widely used for benchmarking NLP tasks. A 500-sample subset was extracted to simulate small long-text scenarios. These datasets allow for a comprehensive evaluation of topic modeling and clustering techniques across varying text lengths (short vs. long) and dataset sizes (large vs. small), providing insights into how different approaches perform under diverse conditions.

\begin{algorithm}[tb]
    \SetAlgoLined
    \caption{Keyword Process}
    \label{algKeywordProcess}
    \textbf{Input:} \\
     $K = \{k_1, k_2, ..., k_{202}\}$ \textcolor{gray}{// keyword list} \\
     $C = \{C_{\text{full}}, C_{\text{short}}, C_{\text{prefix}}, C_{\text{jargon}}\}$ \textcolor{gray}{// keyword categories}

    \For{$w \in C_{\text{short}}$}{
        \If{w is ambiguous}{
            Refine $w$ to full, meaningful form
        }
    }
    \For{$k \in C_{\text{short}}$ \textbf{or} $C_{\text{full}}$}{
        \textcolor{gray}{// generate all POS forms} \\
        $K_{\text{pos}} \gets Wf(k) = \{N(k), V(k), A(k), R(k)\}$ \\
        $K_{\text{lemma}} \gets$ lemmatize($K_{\text{pos}}$)
    }
\end{algorithm}

\subsection{Keyword Processing}\label{sec-keyword-process}

In this work, we utilize a carefully curated list of domain-specific keywords to filter patient feedback into complaint and non-complaint categories. These keywords capture subtle indicators of dissatisfaction, such as legal, medical, or emotional terms that are often overlooked by general-purpose models. We grouped the keywords into four categories: full, shortened, prefix, and jargon. These categories, along with representative examples, are shown in Table~\ref{keyword-list-table}. For instance, a comment like \textit{``I saw another patient’s information on the clinic’s screen, which seems like a HIPAA issue''} highlights a potential privacy concern without explicitly using negative sentiment words, which domain-specific filtering can effectively capture.

A key challenge in keyword processing is handling ambiguous terms, especially shortened forms. For example, the abbreviation compl could represent either complaint or complete, but only the former is relevant in our context. To avoid such ambiguity, we refined the list to include only unambiguous and contextually meaningful terms. Another complexity arises from morphological variations of keywords. Words like argue may appear as argues, argued, or arguing in the dataset. Addressing this requires effective lemmatization, which reduces words to their base forms. However, the success of lemmatization depends heavily on accurate part-of-speech (POS) tagging. Most POS taggers are trained on full sentences rather than isolated tokens, making context-sensitive lemmatization particularly challenging \cite{bergmanis2018context}.

To manage these linguistic nuances, we employ the word\_forms Python library. This tool provides comprehensive mappings of words across POS categories—nouns, verbs, adverbs, and adjectives—allowing us to robustly identify and process all morphological variants of keywords in our dataset. The complete keyword processing procedure is detailed in Algorithm \ref{algKeywordProcess}.

\begin{algorithm}[tb]
    \SetAlgoLined
    \caption{Feedback Process and Filter}
    \label{algFeedbackProcess}
    \textbf{Input:} \\
    $F = \{f_1, f_2, ..., f_{439}\}$ \textcolor{gray}{// patient feedback} \\
    $S = \{a, \text{and}, \text{the}, \text{of}, \ldots\}$ \textcolor{gray}{// stop words} \\
    $K_{\text{lemma}} \gets$ \text{from Algorithm \ref{algKeywordProcess}}

    \For{$f \in F$}{
        $T_f \gets$ tokenize($f$) \\
        $R_f \gets$ remove $T_f$ in ($S$) \\
        \For{$t \in T_f$}{
            \If{$t$ starts with $C_{\text{prefix}}$ \textbf{or} $t \in C_{\text{jargon}}$ \textbf{or} $t \in K_{\text{lemma}}$}{
                $F_{\text{neg}} \gets f$
            }
            \Else{
                $F_{\text{pos}} \gets f$
            }
        }
    } 
\end{algorithm}

\subsection{Feedback Preprocessing}\label{sec-feedback-preprocess}

The effectiveness of topic modeling in NLP depends heavily on the quality of input data. To ensure this, our methodology follows a structured preprocessing pipeline. We begin with tokenization, which segments text into individual terms to make it analyzable. Next, we remove stop words such as ``a," ``and," ``the," and ``of," which are frequent but carry little semantic value. This step helps reduce noise and improve data clarity. We use the ``spacy" library for stop word removal due to its robust and efficient NLP capabilities. This preprocessing step is essential for producing cleaner input that enhances the accuracy and interpretability of downstream topic modeling. The detailed steps are outlined in Algorithm~\ref{algFeedbackProcess}.

% \begin{figure}[tbp]
% \centerline{\includegraphics[width=0.30\textwidth]{img/LDA-Blueprint-1.png}}
% \caption{LDA Blueprint (Source:\cite{dieng2020topic} )}
% \label{Fig:LDA-blueprint}
% \end{figure}

\subsection{Topic Identification}
\subsubsection{Latent Dirichlet Allocation (LDA)}
LDA is a generative probabilistic model that uncovers latent topics within a document collection by modeling each document as a mixture of topics, where each topic is a distribution over words \cite{blei2003latent}. It decomposes the document-word matrix into a document-topic and a topic-word matrix, revealing underlying thematic structures. At the core of LDA, topic proportions for each document are drawn from a Dirichlet prior, and words are generated based on topic assignments and word distributions. In our study, we applied LDA to hospital patient feedback to identify latent themes. We experimented with 3 to 20 topic clusters to balance specificity and generality, using the auto setting for Dirichlet priors to adapt topic distributions. Top words per topic were fine-tuned to enhance interpretability.

% \begin{equation}
% a+b=\gamma\label{eq}
% \end{equation}

\subsubsection{Gibbs Sampling Dirichlet Mixture Model (GSDMM)}

Alongside LDA, we also implement the GSDMM for compliant topic identification. The integration of GSDMM is particularly significant given the nature of our data: most survey responses were brief, often comprising 1 to 3 sentences \cite{mazarura2016comparison}. GSDMM is uniquely suited for such short-texts, as it operates under the assumption that a single document predominantly expresses a single topic. This is a notable deviation from LDA's approach, where a document can encompass multiple topics, making LDA more suitable for longer texts \cite{guo2021improved}. The implementation of GSDMM involved modeling each document as a mixture of topics, with a single dominant topic assigned to each document. For a given document $d$, the topic $k$ is assigned based on the conditional probability $P(k∣d)$, which is calculated using Equation \ref{eqGibs}. In this equation, $n_{k,-d}$ represents the number of documents assigned to topic $k$, excluding the current document $d$; $N_{-d}$ is the total number of documents excluding $d$; $K$ is the total number of topics; is the count of word $n_{k,w}$
w in topic k; $n_{k}$ is the total number of words assigned to topic $k$; and $V$ is the vocabulary size. The hyperparameters $\alpha$ and $\beta$ serve as priors for the document-topic and topic-word distributions, respectively, analogous to their roles in LDA. Through iterative sampling, GSDMM converged to a stable distribution of topics across the documents, effectively capturing the dominant theme in each text fragment.

\begin{equation}
P(k|d) \propto \frac{n_{k,-d} + \alpha}{N_{-d} + K\alpha} \times \frac{n_{k,w} + \beta}{n_{k} + V\beta} \label{eqGibs}
\end{equation}

% \begin{figure*}[tbp]
% \centerline{\includegraphics[width=0.95\textwidth]{img/coherence_score.png}}
% \caption{Coherence Score}
% \label{result_coherance_score}
% \end{figure*}

\subsubsection{BERTopic}
BERTopic \cite{grootendorst2022bertopic} is a topic modeling technique that combines transformer-based embeddings, dimensionality reduction, clustering, and topic representation to extract insights from text. It uses models like BERT or RoBERTa to generate contextual embeddings, which are reduced in dimensionality using UMAP for efficient clustering. HDBSCAN is then applied to group similar texts and determine the number of topics based on data density. BERTopic also allows users to set a fixed number of topics manually. For interpretability, it employs class-based TF-IDF (c-TF-IDF) to extract representative words per cluster. While scalable and flexible, BERTopic assumes each document belongs to a single topic, which may limit its effectiveness on multi-topic texts.

\subsubsection{k-means clustering with BERT embedding (kBERT)}
To address the unique challenges posed by our small dataset of short-texts, we developed a customized topic modeling pipeline that integrates a customized pre-processing approach, BERT embeddings, and k-means clustering. BERT, a transformer-based model pre-trained on a large corpus, generates rich contextual embeddings that capture subtle linguistic nuances and effectively handle Out-of-Vocabulary (OOV) terms, particularly useful for domain-specific language. We only utilized the embedding layer of BERT while freezing all other layers. This ensured that the model acted purely as a feature extractor, preventing any modifications to its pre-trained weights. As a result, there is no risk of overfitting or catastrophic forgetting, making our approach particularly well-suited for small datasets \cite{liu2020survey}. We avoid dimensionality reduction methods, as they can remove key linguistic features, especially in low-resource settings where every word holds significant meaning \cite{raunak2019effective}. The high-dimensional embeddings generated by BERT are preserved to retain the full contextual representation.

Our approach began with careful preprocessing of the textual data, including stopword removal and lemmatization for consistency. We also extended the stopword list with domain-specific terms to improve data quality. Each comment was then encoded using the ‘bert-base-uncased’\footnote{https://huggingface.co/google-bert/bert-base-uncased} model, selected for its balance of performance and generalization. Using the uncased variant helped reduce case-related variability, which is beneficial for short texts. The resulting embeddings were clustered using k-means, with various cluster sizes ($n$ = 3, 5, 10, etc.) tested to find optimal groupings. For interpretability, we identified representative comments closest to each cluster centroid and extracted the most frequent words from them. These words formed the topic labels, ensuring alignment with the semantic meaning of each cluster. This iterative process refined topic selection and enhanced cluster coherence.

\section{Result Analysis}

\subsection{Feedback Filtering}
Our dataset, encompassing 439 comments, underwent keyword matching, which served as a quantitative measure of the filtering process's effectiveness. Notably, the matching frequency of keywords within the comments indicates a substantial impact on the accuracy of the filtering mechanism. Initial results without the use of lemmatization presented 89 keyword matches. The introduction of lemmatization, a process facilitating the reduction of words to their base or `lemma' form, slightly increased this match count to 92. A further improvement is achieved by generating and matching keywords across different POS after lemmatization. This approach propelled the match count to a peak of 102 instances. This increment underscores the pivotal role of nuanced linguistic techniques in enhancing the sensitivity of our keyword-based feedback-filtering approach.

\begin{table}[tb]
  \caption{Coherence Score ($C\textsubscript{v}$) on Large Datasets}
  \label{table-coherence-larrge}
  \centering
  \begin{tabular}{llccccc}
    \toprule
    \multirow{2}{*}{Dataset} & \multirow{2}{*}{Model} & \multicolumn{5}{c}{Number of Topics}  \\
    \cmidrule(lr){3-7}
    & & 5 & 10 & 20 & 30 & 40 \\
    \midrule
    \multirow{4}{*}{Clickbait-title} & LDA & 0.35 & 0.37 & 0.35 & 0.34 & 0.28 \\
      & GSDMM & 0.44 & 0.47 & \textbf{0.49} & 0.48 & 0.47 \\
      & BERTopic$^{\mathrm{a}}$ & 0.38 & 0.46 & 0.44 & 0.43 & 0.46 \\
      & kBERT & 0.40 & 0.39 & 0.41 & \textbf{0.49} & 0.48 \\
      \midrule
      \multirow{4}{*}{20Newsgroup} & LDA & 0.57 & 0.51 & 0.55 &  0.52 & 0.54 \\
      & GSDMM & 0.46 & 0.49 & 0.57 & 0.53 & 0.58 \\
      & BERTopic$^{\mathrm{b}}$ & 0.41 & 0.38 & 0.39 & 0.41 & 0.43 \\
      & kBERT & 0.36 & 0.39 & 0.40 & 0.47  & 0.43 \\
    \bottomrule
  \end{tabular}
  \flushleft
  {\footnotesize $^{\mathrm{a}}$ generated 178 topics ($C_v$=0.48), $^{\mathrm{b}}$ generated 163 topics ($C_v$=0.60)}
\end{table}

\begin{table}[tb]
  \caption{Coherence Score ($C\textsubscript{v}$) on Small Datasets}
  \label{table-coherence-small}
  \centering
  \begin{tabular}{llccccc}
    \toprule
    \multirow{2}{*}{\shortstack{Dataset}} & \multirow{2}{*}{Model} & \multicolumn{5}{c}{Number of Topics}   \\
    \cmidrule(lr){3-7}
    & & 3 & 5 & 10 & 15 & 20  \\
    \midrule
    
    \multirow{4}{*}{\shortstack{Clickbait-title \\ (500)}} & LDA & 0.29 & 0.30 & 0.31 & 0.29 & 0.30  \\
      & GSDMM & 0.28 & 0.38 & 0.32 & 0.37 & 0.36 \\
      & BERTopic$^{\mathrm{a}}$ & N/A & N/A & N/A & N/A & N/A \\
      & kBERT & 0.36 & 0.45 & \textbf{0.45} & 0.44 & 0.42 \\
      \midrule
    \multirow{4}{*}{\shortstack{20Newsgroup\\(500)}} & LDA & 0.53 & 0.49 & 0.47 & 0.49 & 0.44 \\
      & GSDMM & 0.48 & 0.47 & 0.43 & 0.47 & 0.36 \\
      & BERTopic$^{\mathrm{b}}$ & N/A & N/A & N/A & N/A & N/A \\
      & kBERT & 0.41 & 0.48 & 0.43 & 0.43 & 0.41 \\
      \midrule
    \multirow{4}{*}{\shortstack{Patient \\ Feedback\\(439)}} & LDA & 0.34 & 0.35 & 0.34 & 0.38 & 0.36 \\
      & GSDMM & 0.49 & 0.43 & 0.42  & 0.46 & 0.46 \\
      & BERTopic & 0.46 & N/A & N/A & N/A & N/A \\
      & kBERT & \textbf{0.53} & 0.49  & 0.48 & 0.48 & 0.41   \\
    \bottomrule
  \end{tabular}
  \flushleft
  {\scriptsize $^{\mathrm{a}}$ generated 2 topics ($C_v$=0.40), $^{\mathrm{b}}$ generated 2 topics ($C_v$=0.44)}
\end{table}

\subsection{Topic Modeling}
We aimed to balance coherence and clarity in topic modeling. To evaluate performance, we used two metrics: Coherence score ($C_v$) for interpretability and Rank-Biased Overlap (RBO) for topic diversity.

\textbf{Coherence score} ($C_v$) quantitatively measures how semantically related the top words in a topic are. A higher $C_v$ indicates better topic quality, suggesting that the most representative words frequently co-occur and are contextually meaningful. The $C_v$ score is derived from Normalized Pointwise Mutual Information (NPMI) \cite{bouma2009normalized}, which considers the joint and individual probabilities of word co-occurrence. NPMI is defined in Equation~\ref{Eq:NPMI}, where $\epsilon$ prevents division by zero. The final $C_v$ score is calculated as shown in Equation~\ref{Eq:Cv}, where $\gamma$ adjusts the weight given to high-PMI word pairs.

\begin{equation}
    \operatorname{NPMI}\left(w_i^r, w_i^s\right)=\frac{\log _2 \frac{P\left(w_i^r, w_i^s\right)+\epsilon}{P\left(w_i^r\right) P\left(w_i^s\right)}}{-\log _2\left(P\left(w_i^r, w_i^s\right)+\epsilon\right)}
    \label{Eq:NPMI}
\end{equation}

\begin{equation}
    \mathrm{C}_{\mathrm{V}}\left(w_i^r, w_i^s\right)=\mathrm{NPMI}^\gamma\left(w_i^r, w_i^s\right)
    \label{Eq:Cv}
\end{equation}

To assess model effectiveness, we computed the $C_v$ coherence score for each method across three datasets: Clickbait-title, 20Newsgroup, and Patient Feedback. These were split into large and small subsets to test robustness at different data scales. As Patient Feedback consists of short-text comments, we included public short-text datasets for validation. Coherence was measured using the top 10 topic words for long texts and the top 5 for short texts. Best-performing short-text results are highlighted in Tables~\ref{table-coherence-larrge} and \ref{table-coherence-small}.

For large datasets, we evaluated models across multiple topic counts (5 to 40). As shown in Table~\ref{table-coherence-larrge}, GSDMM consistently achieved high coherence on short-text data, scoring 0.49 for 20 topics in Clickbait-title. kBERT also performed well, reaching a score of 0.49 at 30 topics, highlighting its strength in small, short-text scenarios. In contrast, LDA struggled with short texts but performed better on long documents. BERTopic, which automatically determines the number of topics, generated 163 topics in 20Newsgroup ($C_v$ = 0.60) and 178 in Clickbait-title ($C_v$ = 0.49), showing strong performance on large datasets. However, its coherence dropped in smaller datasets, producing only 3 topics ($C_v$ = 0.46) in Patient Feedback.

For small datasets, we generated 3, 5, 10, 15, and 20 topics. As shown in Table~\ref{table-coherence-small}, kBERT achieved the highest coherence score of 0.53 for 3 topics on Patient Feedback, demonstrating its effectiveness in short and sparse data. It also performed well on the Clickbait-title (500) dataset, scoring 0.45 for 5 topics, reinforcing the usefulness of BERT embeddings in low-resource short-text scenarios.

\textbf{Rank-Biased Overlap (RBO)} is a similarity measure for comparing ranked lists, placing more emphasis on top-ranked items and handling incomplete or non-conjoint lists \cite{webber2010similarity}. It is particularly useful for evaluating topic diversity by assessing overlap among top topic words. RBO ranges from 0 (no overlap) to 1 (identical lists), with the parameter $p$ controlling the depth of comparison. To better capture topic diversity, we use the inverted RBO $(1 - \operatorname{RBO})$, where higher values indicate more distinct topics. We compute the average inverted RBO across all topic pairs, as shown in Equations~\ref{Eq:RBO} and \ref{Eq:Inverted_RBO_Avg}.

\begin{equation}
    \operatorname{RBO}(S, T, p) = (1 - p) \sum_{d=1}^{\infty} p^{d-1} \cdot A_d
    \label{Eq:RBO}
\end{equation}

\begin{equation}
    \begin{split}
        \operatorname{IRBO_{avg}}(A, B, C, p) = \frac{1}{3} \Big[ 
        (1 - \operatorname{RBO}(A, B, p)) + \\
        (1 - \operatorname{RBO}(A, C, p)) + \\
        (1 - \operatorname{RBO}(B, C, p)) 
        \Big]
    \end{split}
    \label{Eq:Inverted_RBO_Avg}
\end{equation}

\begin{table}[tb]
\caption{ IRBO\textsubscript{avg} Score on Top 5 words and 3 Topics}
  \label{table-irbo}
  \centering
  \begin{tabular}{lll}
    \toprule
     Dataset & Model  & Score\\
    \midrule
    \multirow{4}{*}{\shortstack{Patient \\ Feedback }} & LDA &  0.83 \\
    & GSDMM &  0.97 \\
    & BERTopic & 1.00  \\
    & kBERT &  1.00 \\
    \bottomrule
  \end{tabular}
\end{table}

We evaluated topic diversity on the Patient Feedback dataset using Average Inverted Rank-Biased Overlap ($\operatorname{IRBO_{avg}}$), where higher scores indicate more distinct topics. As shown in Table~\ref{table-irbo}, kBERT and BERTopic achieved perfect diversity scores of 1.00, while GSDMM scored 0.97 and LDA 0.83, indicating some topic overlap in LDA's results.

\section{Discussion}

Our topic modeling analysis identified three primary areas of concern in patient feedback: `Staff Conduct and Cleanliness', `Safety Protocols Compliance', and `Efficiency of Service'. Positive feedback highlighted professionalism and hygiene practices, reinforcing confidence in staff attentiveness. Similarly, the emphasis on safety measures, particularly mask compliance, indicated patient reassurance regarding institutional health protocols. However, the most significant complaint centered on `Efficiency of Service', with frequent mentions of long wait times and appointment delays, emphasizing the need for healthcare facilities to enhance operational efficiency. These key areas of focus are summarized in Table \ref{topic-name-table}, which presents the dominant topics along with representative terms and examples. A crucial step in our approach was keyword filtering, which effectively differentiated complaints from non-complaints within the dataset. By applying a targeted keyword-based filtration method, we identified 102 complaint-related sentences out of 439 total comments.

\begin{table}[tb]
  \caption{Topics from kBERT with Examples}
  \label{topic-name-table}
  \centering
  \begin{tabular}{p{2.5cm}p{4cm}l}
    \toprule
     Topic words & Example & Complain?\\
    \midrule
    good, clean, great, everyone, professional & Everyone was careful in the emergency department & No\\
     \midrule
     mask, staff, wearing, safe, one & The staff's masks were reassuring in these uncertain times. & No\\
      \midrule
    doctor, time, appointment, waiting, room & Long wait from the time I entered the room until the doctor arrived & Yes\\
    \bottomrule
  \end{tabular}
\end{table}

In evaluating the topic modeling approaches, kBERT consistently outperformed other models in short-text complaint analysis, achieving the highest coherence score ($C_v$ = 0.53) and perfect topic diversity (IRBO = 1.00), indicating clear and well-separated topics. This highlights the strength of BERT embeddings in capturing the nuances of patient feedback. GSDMM also performed well, benefiting from its ability to assign a dominant topic per document. In contrast, BERTopic, while effective on large datasets, showed limited performance on smaller ones ($C_v$ = 0.46). LDA performed better on long texts but struggled with short, sparse data. These results underscore the value of context-aware models like kBERT in healthcare settings. Combined with keyword filtering, this approach enables accurate identification of patient concerns, supporting data-driven service improvements and enhanced care delivery.

\section{Conclusion}
This study explored unsupervised techniques for complaint topic identification in short-text patient feedback, comparing traditional topic modeling methods (LDA, GSDMM) with advanced neural embedding-based approaches (BERTopic, kBERT). Our results demonstrated that kBERT outperforms other models in both coherence and diversity metrics, making it the most effective method for analyzing short and sparse healthcare-related texts. Additionally, a keyword-based filtering approach significantly improved complaint detection. The identified complaint categories include staff conduct and cleanliness, safety protocol compliance, and efficiency of service, and provide actionable insights for healthcare service improvement. 
%In the future, we will also explore few-shot and instruction-based prompting to improve complaint classification and topic extraction in healthcare feedback analysis. To achieve this, we will use open-source LLMs (e.g., LLaMA) locally with sufficient computational resources, preventing data leakage and ensuring compliance.
% In the future, we will use open-source LLMs (e.g., LLaMA) locally to prevent data leakage and ensure compliance, leveraging sufficient computational resources. We will also explore few-shot and instruction-based prompting to improve complaint classification and topic extraction in healthcare feedback analysis.

\section*{Acknowledgment}
We gratefully acknowledge the support of the Advancing a Healthier Wisconsin (AHW) for funding this project.
The authors used AI tools in this manuscript, including Grammarly for grammar checking and ChatGPT for rephrasing and refining, with careful review to ensure contextual accuracy.
%The authors acknowledge the use of AI tools in this manuscript. `Grammarly' was used for grammar checking, and `ChatGPT' assisted in rephrasing and refining sentences, which were carefully reviewed to ensure contextual accuracy.

\bibliographystyle{IEEEtran}
\bibliography{Ref}
\vspace{12pt}

\end{document}